\documentclass[11pt]{article}

\usepackage[final]{acl}

\usepackage{times}
\usepackage{latexsym}
\usepackage{hyperref}
\usepackage{url}
\usepackage{bm}
\usepackage{amsthm,amsmath,amssymb}
\usepackage{colortbl}
\usepackage{multicol}
\usepackage{multirow}
\usepackage{adjustbox}
\usepackage{enumitem}
\usepackage[most]{tcolorbox}
\usepackage{booktabs}
\usepackage{wrapfig}
\usepackage{caption}
\usepackage{marvosym}
\usepackage{titletoc}

\usepackage[T1]{fontenc}

\usepackage[utf8]{inputenc}

\usepackage{microtype}

\usepackage{inconsolata}

\usepackage{graphicx}

\definecolor{darkgrey}{rgb}{0.53,0.53,0.53}
\definecolor{mygrey}{rgb}{0.9,0.9,0.9}

\title{Native Hybrid Attention for Efficient Sequence Modeling}

\author{
 \textbf{Jusen Du\textsuperscript{1,2*}},
 \textbf{Jiaxi Hu\textsuperscript{3}},
 \textbf{Tao Zhang\textsuperscript{1\dag}},
 \textbf{Weigao Sun\textsuperscript{2§\dag}},
 \textbf{Yu Cheng\textsuperscript{4\dag}}
\\
\\
 \textsuperscript{1}Tsinghua University
 \textsuperscript{2}Shanghai AI Laboratory
\\
 \textsuperscript{3}The Hong Kong University of Science and Technology (Guangzhou)
\\
 \textsuperscript{4}The Chinese University of Hong Kong
}

\newcommand\blfootnote[1]{%
\begingroup
\renewcommand\thefootnote{}\footnote{#1}%
\addtocounter{footnote}{-1}%
\endgroup
}

\begin{document}
\maketitle

\blfootnote{\textsuperscript{*}Intern at Shanghai AI Laboratory; \textsuperscript{\dag}Corresponding Authors (taozhang@tsinghua.edu.cn, sunweigao@outlook.com, chengyu@cse.cuhk.edu.hk); \textsuperscript{§}Project Lead.}

\begin{abstract}
Transformers excel at sequence modeling but face quadratic complexity, while linear attention offers improved efficiency but often compromises recall accuracy over long contexts. 
In this work, we introduce \textbf{Native Hybrid Attention (NHA)}, a novel hybrid architecture of linear and full attention that integrates both intra \& inter-layer hybridization into a unified layer design. NHA maintains long-term context in key–value slots updated by a linear RNN, and augments them with short-term tokens from a sliding window. A single \texttt{softmax attention} operation is then applied over all keys and values, enabling per-token and per-head context-dependent weighting without requiring additional fusion parameters. The inter-layer behavior is controlled through a single hyperparameter, the sliding window size, which allows smooth adjustment between purely linear and full attention while keeping all layers structurally uniform. Experimental results show that NHA surpasses Transformers and other hybrid baselines on recall-intensive and commonsense reasoning tasks. Furthermore, pretrained LLMs can be structurally \textbf{hybridized} with NHA, achieving competitive accuracy while delivering significant efficiency gains. Code is available at \url{https://github.com/JusenD/NHA}.

\end{abstract}

\section{Introduction}
Self-Attention \citep{vaswani2017attention} has become the primary architecture for sequence modeling due to its exceptional ability to capture long-term dependencies. However, this power comes at a steep computational cost. The self-attention mechanism has a computational complexity of $O(n^2)$ with respect to the sequence length $n$. This quadratic scaling presents a major obstacle for processing long sequences, a common requirement in domains like long-document analysis \citep{beltagy2020longformer, zaheer2020big} and bioinformatics \citep{dalla2025nucleotide, lin2023evolutionary}. 

\begin{figure*}[h!]
    \centering
    \includegraphics[width=0.9\linewidth]{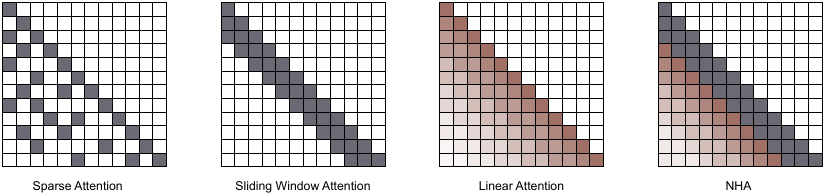}
    \caption{\textbf{Token Dependencies.} Illustration for token dependencies of sparse attention, linear attention, and NHA.}
    \label{fig:attention_map}
    \vspace{-1em}
\end{figure*}

To address this limitation, research community has largely pursued two divergent paths. The first path involves \textbf{sparse attention}. As shown in Fig.~\ref{fig:attention_map}, such methods compute softmax attention over sparsely selected tokens, thereby reducing the computational cost \cite{beltagy2020longformer, jiang2023mistral, lu2025moba, yuan2025native}. Among these approaches, Sliding Window Attention (SWA) currently remains the most efficient implementation, which limits attention computations to a local neighborhood. The second direction explores \textbf{linear sequence modeling}, such as linear attention models \citep{sun2023retentive, yang2024gated, du2025mom, von2025mesanet} and state space models \citep{gu2024mambalineartimesequencemodeling, dao2024transformers, hu2025comba}. These models achieve remarkable $O(n)$ efficiency by compressing the entire sequence history into a fixed-size state, thereby enabling a global receptive field. While SWA cannot capture tokens beyond its local window, the extreme compression of linear models often results in the loss of precise token information. Given that these two approaches have complementary strengths and weaknesses, intra-layer hybrid models \citep{munkhdalai2024leave, lola, lan2025liger} that combine them are a natural next step.

At the same time, a different trade-off motivates another form of hybridization. On one hand, a pure linear model struggles with the theoretical impossibility of perfectly preserving infinite information within a fixed-size memory state. On the other hand, maintaining a full KV cache for every token at every layer of a standard Transformer is not only computationally prohibitive but often unnecessary, as not all layers require the same level of granular global information. This “all-or-nothing" dilemma in information storage across the network depth has motivated the development of inter-layer hybrid models \citep{samba, jamba, zamba, glorioso2024zamba2}.

These two kinds of hybrid architectures can be summarized as follows:
\begin{enumerate}
    \item Intra-layer Hybrid: Typically involving computing the linear attention and local softmax attention separately, then combining them through weighted summation.
    \item Inter-layer Hybrid: Typically involving stacking or alternating different kinds of layers. For instance, stacking 1 softmax attention layer after every 7 linear attention layers to form an inter-layer hybrid architecture.
\end{enumerate}

In this paper, we introduce \textbf{Native Hybrid Attention (NHA)}, \emph{a new hybrid form that natively unifies intra- and inter-layer hybrid architectures into a single, cohesive design}. For intra-layer hybrid, NHA compresses long-term information into a fixed number of slots using a linear RNN model and concatenates them with the local tokens that are inside the window. This combined set of precise, short-term tokens and summarized, long-term slots is then processed by a single, unified \texttt{softmax attention} operation. This enables representing long-term memory in the same $(m\times d)$ key–value slot format as SWA, rather than the outer-product form used in many prior hybrids, ensuring compatibility and efficient concatenation. 

Unlike prior hybrids that compute separate attentions and fuse them through weighted summation, in NHA, the \texttt{softmax attention} mechanism itself learns to dynamically allocate attention between short-term and long-term memory for each query, achieving an automatic and context-aware weighting without the need for manually-tuned or additional parameters. For inter-layer hybrid, all NHA layers share the same design. We can simply change the window size of sliding window attention without modifying the model architecture to control the behavior of each layer. Setting the window size to zero yields a pure linear RNN layer, whereas setting it to the full sequence length recovers a full attention layer. This contrasts with previous inter-layer models that stack different types of layers, which requires managing heterogeneous blocks and their alignment. NHA, therefore, enables flexible, layer-specific configurations without altering the model's fundamental structure.

Our contributions can be summarized as follows:

\begin{itemize}
    \item \textbf{Unified intra- and inter-layer hybrid architecture.} We introduce NHA, which integrates sliding window attention and linear attention within a single, unified \texttt{softmax attention} operation, and achieves inter-layer hybridization simply by adjusting the window size, without changing the model architecture. Furthermore, we develop a chunkwise-parallel Triton kernel for efficient GPU computation.
    \item \textbf{Comprehensive evaluation of hybrid and non-hybrid architectures.} We pretrain and evaluate models from each architectural category. Results demonstrate that hybrid architectures consistently surpass standard Transformers on recall-intensive tasks, with NHA attaining the strongest overall performance.
    \item \textbf{Hybridization of pretrained Transformer models.} We demonstrate that NHA can be applied to pretrained Transformer LLMs. With only brief finetuning, these pretrained models can be adapted into the NHA architecture, achieving competitive performance with improved inference speed.
\end{itemize}

\begin{figure*}[h!]
    \centering
    \vspace{-1.6em}
    \includegraphics[width=0.9\linewidth]{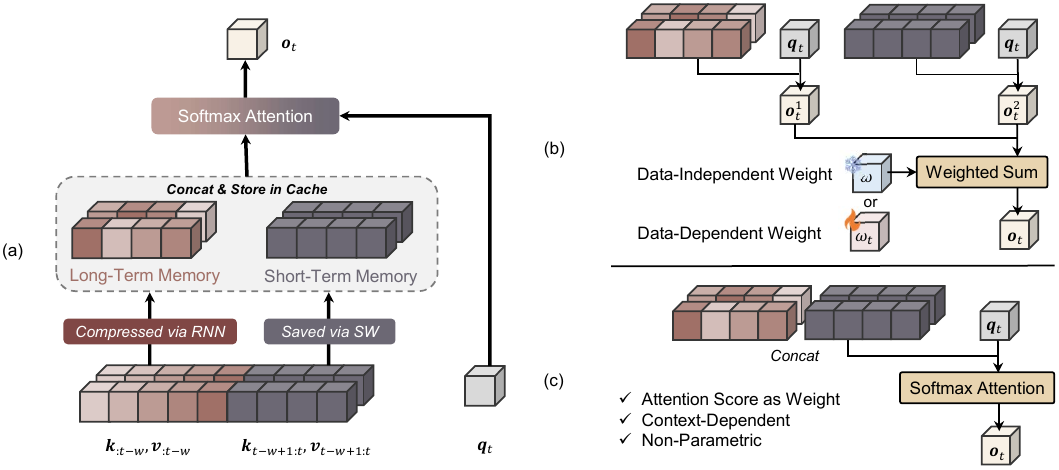}
    \caption{\textbf{Intra-Layer Hybrid in NHA}. (a) NHA compresses historical tokens into fixed-size long-term memory slots (brown) through an RNN update, then concatenates them with recent local context (gray) before applying unified \texttt{softmax attention}. (b) Previous intra-layer hybrid approaches generally compute long-term and short-term outputs separately and combine them through weighted summation. (c) In contrast, NHA employs a non-parametric, context-dependent \texttt{softmax attention} operation to dynamically determine their respective contributions.}
    \label{fig:intra}
\end{figure*}

\section{Preliminary}





To improve the efficiency of Transformers, numerous methods have been developed that restrict the attention window to a fixed number of tokens.

\paragraph{Sliding Window Attention (SWA)}
SWA~\cite{beltagy2020longformer} focuses on a fixed-size window of tokens near the current position, storing these tokens precisely. This approach ensures that immediate context is captured accurately but can miss important information outside the window, limiting its long-term sequence comprehension.

\paragraph{Gated Slot Attention (GSA)}
GSA~\cite{zhang2024gated} compresses the entire sequence into a set of fixed memory slots. Instead of storing precise tokens, GSA uses a gating mechanism to update these slots, blending new and past information as in Eq.~(\ref{eq:long-mem}). This compression allows GSA to maintain a broader context and adaptively manage long-term dependencies. Unlike SWA's localized precision focus, GSA balances memory efficiency with comprehensive sequence understanding.

\section{Naive Hybrid Attention}
\subsection{Level 1: Architectural Hybrid of RNN Memory and Softmax Attention}
Transformer stores the KV cache for the sequence and performs softmax attention over them to compute the output. In contrast, linear attention compresses the entire sequence into a fixed-size matrix memory and replaces the softmax operation with more efficient matrix multiplication. Prior work \cite{peng2021abc, zhang2024gated} has explored hybrid designs that compress past tokens into a small set of memory slots and then apply softmax attention over these slots.
\begin{equation}
\label{eq:long-mem}
\scalebox{0.95}{$
\begin{aligned}
\bm K^{\text{long}}_t &= \text{Diag}(\bm\alpha_t)\, \bm K^{\text{long}}_{t-1} 
   + (1-\bm\alpha_t) \otimes \bm k_t, \\
\bm V^{\text{long}}_t &= \text{Diag}(\bm\alpha_t)\, \bm V^{\text{long}}_{t-1} 
   + (1-\bm\alpha_t) \otimes \bm v_t,
\end{aligned}
$}
\end{equation}
where $t$ represents the current token index, $\bm K^{\text{long}}_t, \bm V^{\text{long}}_t\in \mathbb{R}^{m\times d}$, while $m$ denotes the number of memory slots and $\bm\alpha_t$ denotes an input-dependent tensor of gating value where each element is in the range $[0,1]$.

However, performing softmax attention solely on this compressed memory does not appear to be an ideal solution, as the precise tokens near the current position often play a crucial role. This limitation motivates us to take it a step further by exploring an intra-layer memory hybrid approach.

\subsection{Level 2: Intra-Layer Hybrid}
SWA applies softmax attention within a fixed-size window, focusing on the context of nearby tokens. This approach can be seen as primarily leveraging precise local tokens. The KV cache for tokens within the window remains fixed in size and can be regarded as precise short-term memory.
\begin{equation}
\label{eq:short-mem}
\begin{aligned}
    \bm K^{\text{short}}_t &= \{\bm k^T_{t-w+1}, \bm k^T_{t-w+2}, ..., \bm k^T_t\}^T, \\
    \bm V^{\text{short}}_t &= \{\bm v^T_{t-w+1}, \bm v^T_{t-w+2}, ..., \bm v^T_t\}^T,
\end{aligned}
\end{equation}
where $t$ represents the current token index, $w$ denotes the window size, $\bm K^{\text{short}}_t, \bm V^{\text{short}}_t\in \mathbb{R}^{w\times d}$.

\paragraph{Memory Hybridization} The long-term memory in Eq.~\ref{eq:long-mem} naturally aligns with the short-term memory described above, and the two complement each other effectively. We concatenate them to form a unified, hybrid memory that integrates both long-term and short-term information. Subsequently, the softmax function is used to compute the attention scores, enabling dynamic weighting of long-term and short-term memory.

\begin{equation}
\label{eq:hybrid-mem}
\begin{aligned}
    \bm K^{\text{H}}_t &= \text{Concat}(\bm K^{\text{long}}_t, \bm K^{\text{short}}_t) \in \mathbb{R}^{(m+w) \times d}, \\
    \bm V^{\text{H}}_t &= \text{Concat}(\bm V^{\text{long}}_t, \bm V^{\text{short}}_t) \in \mathbb{R}^{(m+w) \times d}, \\
\end{aligned}
\end{equation}

\begin{equation}
    \bm o_t = \text{softmax}(\frac{\bm q_t \bm (\bm K^H_t)^T}{\sqrt{d}})\bm V^H_t,
\end{equation}
where $t$ represents the current token index, $m$ denotes the number of memory slots, $w$ denotes the window size, and $d$ is the dimension of $\bm q$. 

Token shift ensures that only tokens outside the sliding window update the long-term memory. For $t \le w$, no tokens are removed, and a zero prefix is used to initialize the memory slots. For positional encoding, RoPE is applied within the sliding window, while no positional embedding is added to the long-term memory. Further details and ablation analysis are provided in Appendix~\ref{app:ablation-pos}.

\paragraph{Context-Dependent Fusion}
\label{sec:fusion}
In NHA, the unified softmax over concatenated long-term and short-term memory assigns the proportion of attention to the long-term memory as
\begin{equation}
\label{eq:context-fusion}
\omega_L = \frac{\sum_{i \in \text{long}} \exp(\bm q_t \bm k_i^\intercal)}{\sum_{i \in \text{long}} \exp(\bm q_t \bm k_i^\intercal) + \sum_{j \in \text{short}} \exp(\bm q_t \bm k_j^\intercal)},
\end{equation}
where $\bm q_t$ is the current query, and $\bm k_i, \bm k_j$ are keys from long- and short-term memory respectively. This proportion is computed jointly with token-level weights and depends on similarity scores with all keys in both memory types, thus incorporating information from all preceding tokens.

To further highlight the distinction between unified softmax and output-level weighted fusion, we provide a gradient analysis in Appendix~\ref{app:grad-coupling} and a memory analysis in Appendix~\ref{appendix:mem}. This shows that unified softmax naturally couples gradient flow between long- and short-term memories. We compare the performance of unified softmax fusion and learnable weighted fusion in ablations (Sec.~\ref{sec:ablation}).




Since the window size can vary across different layers in the model, it creates the possibility for an inter-layer hybrid. This flexibility leads us to a deeper exploration of model architecture.

\subsection{Level 3: Inter-Layer Hybrid via Window Size Adjustment}

\begin{figure}[h!]
    \centering
    \vspace{-0.5em}
    \includegraphics[width=0.95\linewidth]{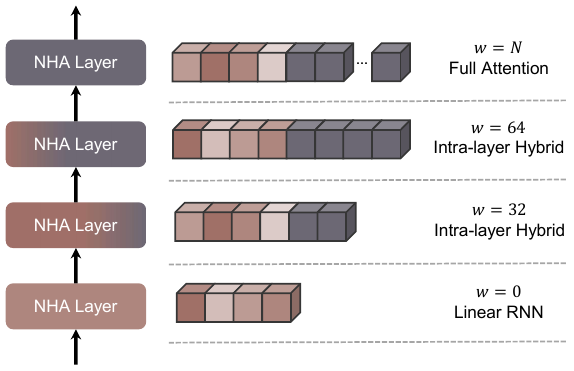}
    \caption{\textbf{Inter-Layer Hybrid in NHA}. All layers share the same NHA architecture. By varying the sliding window size $w$, each layer can behave as full attention ($w=N$), intra-layer hybrid ($w>0$), or pure linear RNN ($w=0$), enabling flexible inter-layer hybridization without changing the model structure.}
    \label{fig:inter}
    \vspace{-0.5em}
\end{figure}

For tasks requiring precise retrieval of long past sequences, pure linear models struggle as it is infeasible to retain all information in a fixed-size memory as the sequence length increases. Storing every $\bm{kv}$ in each layer, as transformers do, appears excessive. This trade-off motivates an inter-layer hybrid approach: strategically placing full attention mechanisms only in select layers, using linear models to \textbf{sparsify} $\bm{kv}$ \textbf{storage across the depth of the Transformer}. This leverages the efficiency of linear models while retaining the robust memory capabilities of attention where most needed. NHA is inherently suited for this hybrid strategy.

The size of the sliding window can vary, ranging from $0$ to the entire sequence length. This flexibility allows us to adjust the "linear ratio" of an NHA layer by modifying the window size. Setting the window size to $0$ reduces the NHA to the pure linear attention model. Conversely, setting the window size to the sequence length transforms the NHA model into a Transformer with a prefix.

NHA enables inter-layer hybridization within a unified architecture. By adjusting the window size in selected layers to span the entire sequence, the model naturally realizes an inter-layer hybrid design without requiring any structural modifications.

\begin{figure}
    \centering
    \includegraphics[width=1.0\linewidth]{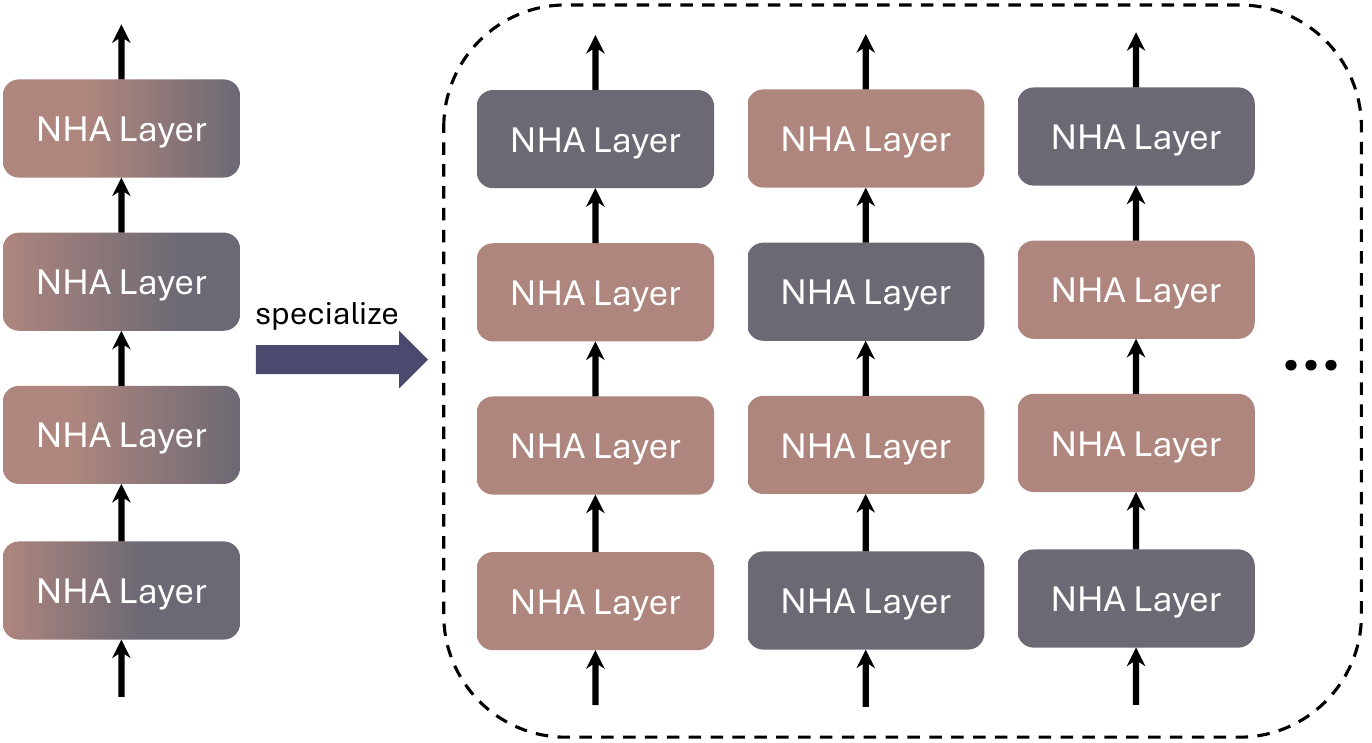}
    \caption{\textbf{NHA Architectural Duality.} A single model trained with randomized windows can be specialized into diverse hybrid configurations for various efficiency-accuracy trade-offs without retraining.}
\end{figure}

\paragraph{Efficient Hybrid Configuration Search.} NHA's architectural uniformity provides a \textit{unified search space} for hybrid configurations. Typically, converting a pretrained Transformer into an efficient hybrid model requires searching for the optimal number and placement of full-attention layers. To search for an optimal hybrid configuration, we need to retrain every hybrid configuration at prohibitive cost. Unlike inter-layer hybrids that force a rigid "0 or 1" choice (either pure Linear or pure Transformer) for each layer, NHA exhibits a unique "duality". This allows the same model to seamlessly switch between high-speed RNN mode and high-precision Full Attention mode during inference without any retraining.  We can efficiently identify the optimal hybrid patterns by simply adjusting the window sizes during inference (Table~\ref{tab:inference_search}). Furthermore, this duality offers a promising path for dynamic deployment, potentially allowing a model to adapt its precision and speed to varying hardware constraints without retraining.

\subsection{Chunkwise Parallel Form}
NHA is able to implement in a chunkwise parallel form. Let the sequence is divided into $N = \lceil T/C \rceil$ chunks of size $C$. For each chunk $[t]$, let the cumulative and reverse gates be $\overrightarrow{\mathcal{A}}_{[t],i} = \prod_{j=1}^{i} \mathbf{A}_{[t],j}$ and $\overleftarrow{\mathcal{A}}_{[t],i} = \prod_{j=i+1}^C \mathbf{A}_{[t],j}$.

\paragraph{Linear Memory Update} We use a gated linear RNN update \cite{yang2023gated}, following the associative structure of GSA \cite{zhang2024gated}:
\begin{equation}
    \mathbf{S}_{[t]} = \mathrm{Diag}(\overrightarrow{\mathcal{A}}_{[t],C})\,\mathbf{S}_{[t-1]} + (\mathbf{K}_{[t]} \odot \overleftarrow{\mathcal{A}}_{[t]})^\top \mathbf{V}_{[t]}.
\end{equation}

\noindent\textbf{Hybrid Attention Logits.} For each chunk, we compute two sets of attention logits:

\begin{enumerate}
    \item \emph{Linear channel logits:}
    \begin{equation}
        \mathbf{O}^k_{[t]} = \overline{\mathbf{Q}}_{[t]} \mathbf{S}_{[t-1]} + (\overline{\mathbf{Q}}_{[t]} \overline{\mathbf{K}}_{[t]}^\top \odot \mathbf{M}_{[t]}) \mathbf{I}_{[t]},
        \label{eq:absorb}
    \end{equation}
    where $\overline{\mathbf{Q}}_{[t]} = \mathbf{Q}_{[t]} \odot \overrightarrow{\mathcal{A}}_{[t]}$, $\overline{\mathbf{K}}_{[t]} = \mathbf{K}_{[t]} \odot (\overleftarrow{\mathcal{A}}_{[t]} / \overrightarrow{\mathcal{A}}_{[t],C})$, $\mathbf{I}_{[t]} = 1 - \mathbf{A}_{[t]}$ and $\mathbf{M}_{[t]}$ is the causal mask.

    \item \emph{Shifted sliding window logits:}
    \begin{equation}
    \mathbf{L}^{\mathrm{swa}}_{[t]} = \mathbf{Q}_{[t]}\,\widetilde{\mathbf{K}}_{\mathrm{win}([t])}^\top + \log \mathbf{M}_{\mathrm{win}},
\end{equation}
    where $\mathrm{win}([t])$ indexes positions in the current chunk and its recent $W$ successor, $\widetilde{\mathbf{K}} = \big[\,\mathbf{0}_W;\,\mathbf{K}\,\big]_{:T}$, $\widetilde{\mathbf{V}} = \big[\,\mathbf{0}_W;\,\mathbf{V}\,\big]_{:T}$.
\end{enumerate}

The two logits are concatenated for softmax:
\begin{equation}
\mathbf{P}^{\mathrm{mix}}_{[t]} = \operatorname{softmax}\!\big([\mathbf{O}^k_{[t]};\,\mathbf{L}^{\mathrm{swa}}_{[t]}]\big).
\end{equation}
The first $m$ columns correspond to the linear channel weights $\mathbf{Q}^v_{[t]}$, and the remainder to the sliding window weights $\mathbf{P}^{\mathrm{swa}}_{[t]}$.

\noindent\textbf{Value Aggregation.} We compute the outpus of linear memory branch and the sliding window branch:
\begin{equation}
\begin{aligned}
\mathbf{O}^v_{[t]} = \overline{\mathbf{Q}}^v_{[t]} \mathbf{S}^{v}_{[t-1]}
&+ (\overline{\mathbf{Q}}^v_{[t]} \mathbf{I}_{[t]}^\top \odot\mathbf{M}_{[t]}) \overline{\mathbf{V}}_{[t]}, \\
\mathbf{O}^{\mathrm{swa}}_{[t]} &= \mathbf{P}^{\mathrm{swa}}_{[t]} \widetilde{\mathbf{V}}_{\mathrm{win}([t])},
\end{aligned}
\end{equation}
where $\overrightarrow{\mathcal{A}}$, $\overleftarrow{\mathcal{A}}$ is absorbed into $\overline{\mathbf{Q}}^v_{[t]}$, $\overline{\mathbf{V}}_{[t]}$ the same as in Eq.~\ref{eq:absorb}, and $\widetilde{\mathbf{V}}$ represents the shifted $\mathbf{V}$.

The final chunkwise output is:
\begin{equation}
    \mathbf{O}_{[t]} = \mathbf{O}^v_{[t]} + \mathbf{O}^{\mathrm{swa}}_{[t]}.
\end{equation}

\subsection{Connections to MesaNet and Atlas}
In the following, we show that our proposed NHA is closely related to recent architectures such as MesaNet \citep{von2025mesanet} and Atlas \cite{behrouz2025atlas}. In particular, when equipped with matrix memory, MesaNet and Atlas can be interpreted as performing recursive least squares computations, where the models are optimized under a global L2 loss, specifically, $\mathcal{L}=\sum_{i=1}^t \left({\alpha_i}\left\|\bm{v}_i-\bm{k}_i\bm{S}\right\|^2 \right)$. In practice, the accumulation is typically confined to a limited range, thereby preserving relatively precise short-term memory while compressing long-term memory. The range aligns with the conjugate gradient step size in MesaNet and the window size in Atlas. 

SWA can be viewed as a Householder-like transformation constructed from unit vectors $\bm{e}_t\in\mathbb{R}^m$. From this perspective, SWA is an unlearnable extreme case of the Delta rule, where the model retains all precise information within the window while discarding everything beyond it. NHA adopts this mechanism to preserve short-term precise memory while leveraging a linear recurrence to compress long-term memory for key and value, respectively. In contrast, the KV updates in MesaNet and Atlas are performed together, which precludes the incorporation of the softmax operation.



\section{Experiments}
\label{sec:experiments}
To validate the design and capabilities of NHA, our experiments are structured to answer the following key research questions (RQs), with concise answers provided in Appendix~\ref{app:ans}.



\begin{tcolorbox}[
  notitle,
  rounded corners,
  colframe=darkgrey,
  colback=gray!10,
  boxrule=1pt,       
  boxsep=0pt,
  left=0.15cm,
  right=0.17cm,
  enhanced,
  shadow={1pt}{-1pt}{0pt}{opacity=0.5,mygrey}, 
  toprule=0.75pt,       
  before skip=0.65em,
  after skip=0.75em
]
\begin{enumerate}[label=\textbf{RQ\arabic*:}, leftmargin=*, labelindent=0pt, itemsep=0.2em]
    \item \textit{How does NHA perform against Transformer and other hybrid models?}  (Sec.~\ref{sec:recall},~\ref{sec:common},~\ref{app:long_context})
    \item \textit{Can NHA achieve competitive performance to standard Transformers while offering lower cost?} (Sec.~\ref{sec:ops_efficiency},~\ref{sec:hybridation})
    \item \textit{How do NHA's hybrid components contribute to performance?} (Sec.~\ref{sec:ablation}, App.~\ref{appendix:mem},~\ref{app:ablation_sw})
    \item \textit{Is NHA scalable for production-level LLMs?}  (Sec.~\ref{sec:hybridation}, App.~\ref{app:qwen},~\ref{app:attn_layers})
\end{enumerate}
\end{tcolorbox}


\subsection{Experiment Setup}
To answer the research questions, we conduct comprehensive experiments using a consistent setup. We pretrain all baseline hybrid models from scratch to ensure fair and consistent comparison. Following standard practice, we adopt a “one Transformer every eight layers” stacking strategy: the 340M-parameter models insert a Transformer layer as the seventh layer in every eight-layer block, while 1.3B-parameter models insert it as the first layer in every block. Comprehensive evaluation is conducted at both 340M and 1.3B scales. Additional implementation details, including dataset and training schedule, are provided in Appendix~\ref{app:exp-details}.


\begin{table*}[t!]
    \centering
    \begin{adjustbox}{width=1.05\textwidth, center}
    \begin{small}
    \setlength{\tabcolsep}{4pt}
    \renewcommand{\arraystretch}{1.2}
    \begin{tabular}{lcc|cccccc>{\columncolor{red!7}}c|cccccc>{\columncolor{red!7}}c}
        \toprule
        \multirow{2}{*}{\textbf{Model}} & \multirow{2}{*}{\shortstack{\textbf{Wiki.} \\ ppl$\downarrow$}} & \multirow{2}{*}{\shortstack{\textbf{LMB.} \\ ppl$\downarrow$}} & \textbf{ARC$_e$} & \textbf{ARC$_c$} & \textbf{PIQA} & \textbf{Hella.} & \textbf{LMB.} & \textbf{Wino.} & \textbf{Avg.} & \textbf{FDA} & \textbf{SWDE} & \textbf{TQA} & \textbf{NQ} & \textbf{SQD} & \textbf{Drop} & \textbf{Avg.} \\
        & & & acc & $\mathrm{acc_n}$ & acc & $\mathrm{acc_n}$ & acc & acc & acc & acc & acc & acc & acc & acc & acc & acc \\
        \midrule
        \rowcolor{gray!15}\multicolumn{17}{l}{\textbf{\textit{340M Params 15B Tokens}}} \\
        Trans++  & 26.88 & 42.15 & 44.91 & \textbf{25.94} & 64.31 & 34.95 & 32.84 & 51.07 & 42.34 & 46.14 & 25.87 & 45.97 & 18.94 & 33.22 & 20.03 & 31.70 \\
        \midrule
        GLA      & 28.78 & 39.00 & 44.53 & 22.27 & 63.93 & 34.84 & 32.27 & 51.38 & 41.54 & 11.26 & 16.78 & 43.90 & 12.77 & 27.85 & 17.68 & 21.71 \\
        GSA      & 28.17 & 42.57 & 45.50 & 24.23 & 64.85 & 35.00 & 30.78 & 50.43 & 41.80 &  6.36 & 16.87 & 42.18 & 14.60 & 21.90 & 16.72 & 19.77 \\
        GDN      & 26.47 & \textbf{31.69} & 46.04 & 23.55 & \textbf{66.05} & 35.18 & 34.35 & 50.83 & 42.67 & 20.53 & 23.24 & 44.91 & 14.98 & 28.55 & 16.48 & 24.78 \\
        \midrule
        Mamba2-H & 25.81 & 43.35 & \textbf{48.23} & 24.49 & 65.23 & \textbf{36.31} & 30.86 & 48.62 & 42.29 & 55.68 & 39.55 & 45.50 & 18.34 & 27.98 & 17.63 & 34.11 \\
        GLA-H    & 27.94 & 40.12 & 45.20 & 24.32 & 63.93 & 34.42 & 31.83 & 49.96 & 41.61 & 58.86 & 39.36 & 44.14 & 17.01 & 31.37 & 18.21 & 34.83 \\
        GSA-H    & 26.19 & 46.05 & 45.54 & 23.72 & 65.23 & 35.19 & 30.91 & 50.99 & 41.93 & \textbf{62.13} & 45.36 & 43.78 & \textbf{20.62} & 31.17 & 18.78 & 36.97 \\
        GDN-H    & \textbf{25.67} & 51.79 & 45.75 & 23.21 & 65.56 & 35.80 & 29.30 & 51.38 & 41.83 & 52.32 & 38.80 & 39.40 & 15.24 & 29.83 & 17.92 & 32.25 \\
        NHA      & 25.97 & 38.38 & 47.69 & 24.32 & 65.67 & 36.03 & 32.72 & \textbf{52.09} & \textbf{43.09} & 53.86 & \textbf{48.55} & \textbf{46.56} & 20.56 & \textbf{38.60} & \textbf{23.48} & \textbf{38.60} \\
        \midrule
        \rowcolor{gray!15}\multicolumn{17}{l}{\textbf{\textit{1.3B Params 100B Tokens}}} \\
        Trans++  & 17.61 & 15.86 & 55.01 & 28.07 & 70.08 & 49.21 & 45.60 & 56.27 & 50.71 & 44.32 & 32.43 & 58.47 & 24.49 & 42.59 & 21.56 & 37.31 \\
        \midrule
        GLA      & 17.61 & 15.38 & 55.18 & 27.56 & 69.86 & 48.89 & 46.05 & 53.91 & 50.24 & 27.61 & 30.93 & 56.28 & 22.27 & 35.04 & 19.45 & 31.93 \\
        GSA      & 16.69 & 12.62 & 58.33 & 28.33 & 72.25 & 50.98 & 47.43 & 53.43 & 51.79 & 23.25 & 32.80 & 57.05 & 22.96 & 35.57 & 20.65 & 32.05 \\
        GDN      & 17.14 & \textbf{11.35} & 56.82 & 27.39 & 71.76 & 49.77 & 49.10 & 51.78 & 51.10 & 30.25 & 27.65 & 58.23 & 23.22 & 34.06 & 20.36 & 32.30 \\
        \midrule
        Mamba2-H & \textbf{16.00} & 12.77 & 57.74 & 28.24 & 71.93 & \textbf{52.73} & 48.46 & 55.49 & 52.43 & 65.30 & 50.89 & 55.92 & 28.54 & 41.99 & 22.46 & 44.18 \\
        GLA-H    & 17.58 & 15.72 & 56.73 & 27.56 & 70.40 & 48.71 & 46.71 & 54.46 & 50.76 & \textbf{69.48} & 46.77 & 57.05 & 27.15 & 40.28 & 21.47 & 43.70 \\
        GSA-H    & 16.22 & 23.09 & 57.79 & 28.07 & 71.71 & 51.70 & 41.92 & 53.35 & 50.76 & 68.30 & 51.83 & 58.59 & 26.42 & 42.36 & 22.42 & 44.99 \\
        GDN-H    & 16.02 & 11.68 & 58.12 & 27.99 & 70.57 & 52.52 & \textbf{49.78} & 56.27 & 52.54 & 66.76 & 49.02 & 59.06 & \textbf{29.30} & 41.15 & 24.01 & 44.88 \\
        NHA      & 16.16 & 12.58 & \textbf{58.71} & \textbf{28.50} & \textbf{72.03} & 52.08 & 49.19 & \textbf{56.83} & \textbf{52.89} & 68.30 & \textbf{52.48} & \textbf{59.60} & 27.18 & \textbf{45.58} & \textbf{25.44} & \textbf{46.43} \\
        \bottomrule
    \end{tabular}
    \end{small}
    \end{adjustbox}
    \caption{\textbf{Results on Common-Sense Reasoning Tasks and Recall-Intensive Tasks.} GDN denotes Gated DeltaNet; “-H” denotes hybrid with Transformer layers.}
    \label{table:combined_results}
\end{table*}

\begin{figure*}
    \begin{adjustbox}{width=1.\linewidth, center}
    \begin{minipage}{0.38\linewidth}
        \centering
        \includegraphics[width=1\columnwidth]{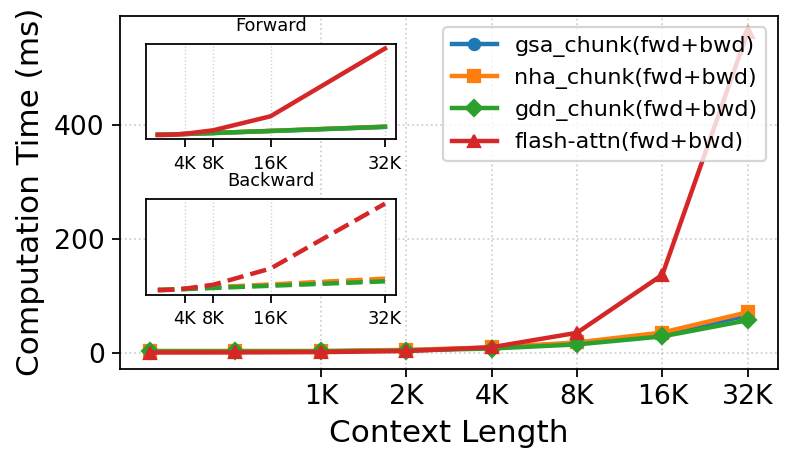}
        \vspace{-2em}
        \caption{\textbf{Operator Training Throughput.} Throughput for forward and backward pass.}
        \label{fig:ops_speed}
    \end{minipage}
    \hspace{4mm}
    \begin{minipage}{0.58\linewidth}
        \begin{adjustbox}{width=1\columnwidth, center}
            \small
            \centering
            \setlength{\tabcolsep}{3pt}
            \renewcommand{\arraystretch}{1.2}
            \begin{tabular}{l|ccc|ccc|ccc|ccc}
            \toprule
            \multirow{2}{*}{\textbf{Model}} & \multicolumn{3}{c|}{\textbf{NIAH-MK}} & \multicolumn{3}{c|}{\textbf{NIAH-MQ}} & \multicolumn{3}{c|}{\textbf{RULER-CWE}} & \multicolumn{3}{c}{\textbf{RULER-Hotpot}} \\
            & 1K & 2K & 4K & 1K & 2K & 4K & 1K & 2K & 4K & 2K & 4K & 8K \\
            \midrule
            Transformer & \textbf{90.2} & \textbf{63.2} & 0.4 & \textbf{81.8} & 64.3 & 13.1 & 42.9 & 20.9 & 0.0 & 21.3 & 2.8 & 0.6 \\
            Mamba2-H     & 13.6 & 3.8  & 4.8 & 52.5 & 55.8 & \textbf{35.0} & 14.5 & 9.0  & 8.5 & 4.4  & 19.4 & 16.0 \\
            GLA-H        & 46.0 & 23.6 & 12.4& 37.9 & 30.1 & 25.0 & 50.4 & 12.7 & 10.4& 26.2 & 20.6 & 19.2 \\
            GSA-H        & 20.0 & 11.2 & 6.4 & 77.7 & \textbf{68.5} & 34.4 & 42.1 & 24.4 & \textbf{15.4}& 24.8 & 20.6 & 18.0 \\
            GDN-H        & 59.6 & 29.4 & 15.0& 56.1 & 41.4 & 27.1 & \textbf{64.7} & 31.0 & 4.1 & 26.5 & 21.6 & 16.8 \\
            NHA          & 81.4 & 51.8 & \textbf{21.6}& 62.7 & 50.6 & 28.4 & 63.8 & \textbf{33.2} & 9.3 & \textbf{27.4} & \textbf{26.0} & \textbf{24.8} \\
            \bottomrule
            \end{tabular}
            \end{adjustbox}
            \captionof{table}{\textbf{Results on RULER tasks.} Models are trained with a context length of 2K.}
            \label{tab:ruler_selected}
    \end{minipage}
    \end{adjustbox}
\end{figure*}

\subsection{Recall-Intensive Tasks}
\label{sec:recall}

Linear models inherently face challenges in recall-intensive tasks due to their fixed-size memory states, which often leads to an performance gap when compared to Transformers. To evaluate NHA's capability in balancing long-term memory with efficient retrieval, we conducted experiments on six recall-intensive benchmarks including FDA \citep{arora2023language}, SWDE \citep{arora2023language, lockard-etal-2019-openceres}, SQuAD \citep{rajpurkar2018know}, NQ \citep{kwiatkowski2019natural}, TriviaQA \citep{joshi2017triviaqa} and Drop \citep{dua2019drop}. We compared NHA's performance against various pure linear models, standard Transformers, and other hybrid architectures, as summarized in Table~\ref{table:combined_results}. Our results demonstrate that hybrid models, in general, achieve significantly better performance than Transformer models with only a limited number of full attention layers and NHA consistently achieves the best results across these benchmarks.

\subsection{Commonsense Reasoning Tasks}
\label{sec:common}

Commonsense reasoning benchmarks evaluate a broader and more fundamental set of capabilities, such as semantic understanding, and general world knowledge. We evaluated NHA on a suite of widely-recognized commonsense reasoning tasks, including WikiText \citep{merity2016pointer}, LAMBADA \citep{paperno2016lambada}, ARC-Easy, ARC-Challenge \citep{Clark2018ThinkYH}, HellaSwag \citep{zellers2019hellaswag}, PiQA \citep{Bisk2020} and WinoGrande \citep{sakaguchi2019winogrande}. All evaluations were performed using the lm-eval-harness framework \citep{eval-harness}. As shown in Table~\ref{table:combined_results}, NHA achieves the highest average score across these tasks, demonstrating that its hybrid design effectively preserves strong general reasoning abilities while enhancing efficiency.

\subsection{Long-Context Benchmarks}
\label{app:long_context}
As shown in Table~\ref{tab:ruler_selected}, we evaluate models on tasks from the RULER \cite{hsieh2024ruler} benchmark, including NIAH-MK, NIAH-MQ, RULER-CWE, and RULER-Hotpot. The models are trained with a context length of 2K and tested on extrapolation up to 8K. While Transformers still hold advantages on in-length needle-in-a-haystack tasks, NHA exhibits stronger extrapolation and achieves the best results across multiple tasks.


\begin{table*}[h!]
    \centering
    \begin{adjustbox}{width=1.05\textwidth, center}
    \begin{small}
    \renewcommand{\arraystretch}{1.3} 
    \renewcommand{\multirowsetup}{\centering}
    \setlength{\tabcolsep}{2.5pt}
    \begin{tabular}{lc|ccccccc>{\columncolor{red!7}}c|cccccc>{\columncolor{red!7}}c}
        \toprule
         \multirow{2}{*}{Model \& Scale} & \multirow{2}{*}{\textbf{\shortstack{Training \\ Tokens (B)}}} & ARC$_e$ & ARC$_c$ & Hella. & LMB. & PIQA & Wino. & MMLU & \multirow{1}{*}{Avg.} & FDA & SWDE & SQD. & NQ & TQA.  & Drop  & \multirow{1}{*}{Avg.}\\
         & & acc & acc\rlap{$_{\text{n}}$} & acc\rlap{$_{\text{n}}$} & acc & acc & acc & acc & acc & acc & acc & acc & acc & acc & acc & acc\\
        \midrule
        \rowcolor{gray!15}\multicolumn{17}{l}{\textbf{\textit{Transformer}}} \\
        Mistral-7B & 8000 & 80.85 & 54.01 & 81.08 & 69.49 & 80.90 & 73.64 & 62.50 & 71.78 & 75.30 & 65.32 & 54.01 & 42.76 & 73.16 & 27.55 & 56.35 \\
        Qwen3-8B & 36000 & 83.42 & 56.74 & 74.90 & 61.11 & 76.61 & 68.11 & 74.93 & 70.83 & 73.48 & 76.85 & 59.89 & 45.01 & 72.63 & 36.61 & 60.75 \\
        \rowcolor{blue!7} Qwen2.5-7B & 18000 & 80.43 & 51.45 & 78.93 & 64.70 & 78.78 & 73.09 & 74.18 & 71.65 & 82.38 & 75.45 & 54.28 & 48.43 & 77.49 & 38.28 & 62.72 \\
        \rowcolor{yellow!7} Llama-3-8B & 15000 & 80.30 & 53.07 & 79.14 & 68.89 & 79.60 & 72.77 & 65.34 & 71.30 & 82.92 & 72.07 & 54.22 & 43.24 & 74.17 & 33.88 & 60.08 \\
        \midrule
        \rowcolor{gray!15}\multicolumn{17}{l}{\textbf{\textit{Linear/Subquadratic}}} \\
        Mamba-7B & 1200 & 77.61 & 46.84 & 77.93 & 65.53 & 79.87 & 71.74 & 33.19 & 64.67 & 37.06 & 43.30 & 46.93 & 33.39 & 71.56 & 25.30 & 42.92 \\
        FalconMamba-7B & 5500 & 83.33 & 58.45 & 80.22 & 61.60 & 80.20 & 75.53 & 60.27 & 71.37 & 48.32 & 59.51 & 49.31 & 37.85 & 75.06 & 30.09 & 50.02 \\
        RWKV-6-World-7B & 1420 & 73.61 & 43.86 & 75.17 & 69.26 & 78.35 & 67.72 & 43.39 & 64.48 & 56.86 & 51.55 & 46.62 & 37.03 & 69.85 & 28.70 & 48.44 \\
        Griffin-7B & 300 & 75.40 & 47.90 & 78.60 & - & 81.00 & 72.60 & 39.30 & - & - & - & - & - & - & - & - \\
        \midrule
        \rowcolor{gray!15}\multicolumn{17}{l}{\textbf{\textit{Hybrid}}} \\
        StripedHyena-7B$_{\text{(16)}}$ & - & 78.62 & 53.75 & 79.11 & 60.78 & 80.03 & 70.32 & 54.07 & 68.10 & 75.30 & 69.82 & 53.81 & 42.41 & 71.86 & 32.34 & 57.59 \\
        Zamba-7B$_{\text{(13)}}$ & 1000 & 71.72 & 45.56 & 80.81 & 64.84 & 80.41 & 72.14 & 57.68 & 67.59 & 76.02 & 71.70 & 50.69 & 42.10 & 70.56 & 25.78 & 56.14 \\
        Zamba2-7B$_{\text{(9)}}$ & 2100 & 80.13 & 56.23 & 81.52 & 59.09 & 79.05 & 77.19 & 67.27 & 71.50 & 71.48 & 64.48 & 50.15 & 40.13 & 71.86 & 29.23 & 54.56 \\
        \midrule
        \rowcolor{blue!7} NHA-Qwen2.5-7B$_{\text{(4)}}$ & 10 & 82.74 & 55.97 & 76.49 & 64.54 & 78.73 & 71.03 & 68.80 & 71.19 & 62.58 & 44.61 & 47.26 & 37.47 & 71.98 & 37.47 & 50.23 \\
        \rowcolor{yellow!7} NHA-Llama-3-8B$_{\text{(4)}}$ & 10 & 80.76 & 52.47 & 78.93 & 67.26 & 79.71 & 72.85 & 60.21 & 70.31 & 73.84 & 67.67 & 55.16 & 42.10 & 73.05 & 34.02 & 57.64 \\
        \bottomrule
    \end{tabular}
    \label{tab:main_results}
    \end{small}
    \end{adjustbox}
    \vspace{-0.5em}
    \caption{\textbf{Comparison of Pretrained LLMs and Their NHA Hybrids.} Performance of original Transformer, linear/subquadratic, and hybrid baselines versus our NHA-hybridized Llama-3-8B and Qwen2.5-7B. Hybrid models show the number of full-attention layers in subscript parentheses.}
    \label{table:finetune}
\end{table*}

\begin{table*}[h!]
    \begin{adjustbox}{width=1.05\textwidth, center}
    \begin{small}
    \renewcommand{\arraystretch}{1.3} 
    \renewcommand{\multirowsetup}{\centering}
    \setlength{\tabcolsep}{2.5pt}
    \begin{tabular}{l|cccccc>{\columncolor{red!7}}c|ccccc>{\columncolor{red!7}}c}
    \toprule
        \textbf{Model} & \textbf{ARC$_e$} & \textbf{ARC$_c$} & \textbf{Hella.} & \textbf{LMB.} & \textbf{PIQA} & \textbf{Wino.} & Avg. & \textbf{GPQA} & \textbf{IFVL.} & \textbf{HEval.} & \textbf{MaQA.} & \textbf{MMLU} & Avg. \\
    \midrule
        Qwen3-30BA3B & 78.87 & \textbf{56.57} & 77.63 & 63.05 & 79.49 & 69.61 & 70.87 & 40.62 & 30.58 & 15.85 & 58.09 & \textbf{77.84} & 44.60 \\
        NHA-Qwen3-30BA3B & \textbf{82.24} & 56.40 & \textbf{79.10} & \textbf{68.85} & \textbf{80.96} & \textbf{73.80} & \textbf{73.56} & \textbf{41.96} & \textbf{30.94} & \textbf{26.22} & \textbf{61.64} & 75.51 & \textbf{47.25} \\
    \bottomrule
    \end{tabular}
    \end{small}
    \end{adjustbox}
    \caption{\textbf{Performance Comparison.} Qwen3-30B-A3B versus NHA-Qwen3-30B-A3B on standard benchmarks.}
    \label{tab:qwen}
    \vspace{-1em}
\end{table*}

\subsection{Operator Efficiency}
\label{sec:ops_efficiency}


We benchmark the forward and backward computation time of different attention operators using the \texttt{Triton-Testing-Benchmark} \citep{tillet2019triton} on one NVIDIA H100-80G GPU. For NHA, we set the long-term memory slot size to 32 and the sliding window size to 32. For GSA, we use 64 memory slots. Figure~\ref{fig:ops_speed} reports the results for varying input sequence lengths. FlashAttention is the fastest on short sequences due to its optimized full-attention kernel, but its quadratic complexity causes computation time to grow sharply for long sequences. In contrast, NHA and GSA maintain near-linear scaling, and NHA matches GSA’s speed across all lengths, as its intra-layer hybridization with a small sliding window adds negligible overhead.

\subsection{Pretrained LLM Hybridization}
\label{sec:hybridation}
Having established the fundamental performance and efficiency of NHA on smaller models through pretraining from scratch, we sought to further validate its practical utility and scalability. For this purpose, we conducted an experiment to structurally hybridize pre-trained open-source Transformer LLMs into NHA-based hybrids. This process involves replacing selected full-attention layers with NHA modules to obtain a faster and more efficient model without requiring full retraining.

\subsubsection{Experiment Setup}
We applied NHA hybridization to two pretrained LLMs: Llama-3-8B and Qwen2.5-7B, each configured with 4 full-attention layers within the hybrid architecture. This hybridization is followed by a lightweight finetuning stage, ensuring that our findings are not tied to a single architecture. Detailed initialization, layer configuration, and training procedures are provided in Appendix~\ref{appendix:finetune-setup}. To further test scalability, we scale up to the Qwen3-30B-A3B model with detailed configuration in Appendix~\ref{app:qwen}.

\subsubsection{Results}
\label{sec:hybridation-results}
We evaluated the finetuned NHA-Qwen2.5-7B and NHA-Llama-3-8B against their original models, leading linear models and hybrid architectures (Table~\ref{table:finetune}). NHA-Llama closely matches the original Llama and consistently offers a better trade-off between recall accuracy and computational efficiency than other efficient models. NHA-Qwen2.5 maintains strong performance on commonsense reasoning tasks. The larger gap on recall-intensive tasks may be partly due to the limited 10B-token finetuning budget, the distribution shift between the SlimPajama corpus and Qwen’s pretraining data, and the hardware constraints that limited training context to 2K tokens. Both models also show a drop on MMLU, a benchmark where even state-of-the-art hybrid models underperform. Notably, despite the small finetuning budget, NHA reaches performance comparable to the strongest same-scale hybrid baselines. We further note that for existing hybrid models, the average recall accuracy tends to correlate with the number of full attention layers. Remarkably, our NHA-Llama3-8B achieves the best performance with only 4 full attention layers, highlighting its efficiency advantage (see Fig.~\ref{fig:hybrid}).

Scaling further, hybridizing Qwen3-30B-A3B also yields competitive results across diverse benchmarks (Table~\ref{tab:qwen}, Appendix~\ref{app:qwen}), confirming that NHA can be applied to very large-scale models while reducing reliance on full attention layers.

\subsubsection{Efficiency}
\label{sec:hybridation-efficiency}
\vspace{-0.5em}

\begin{figure}[h!]
  \begin{center}
    \includegraphics[width=0.5\textwidth]{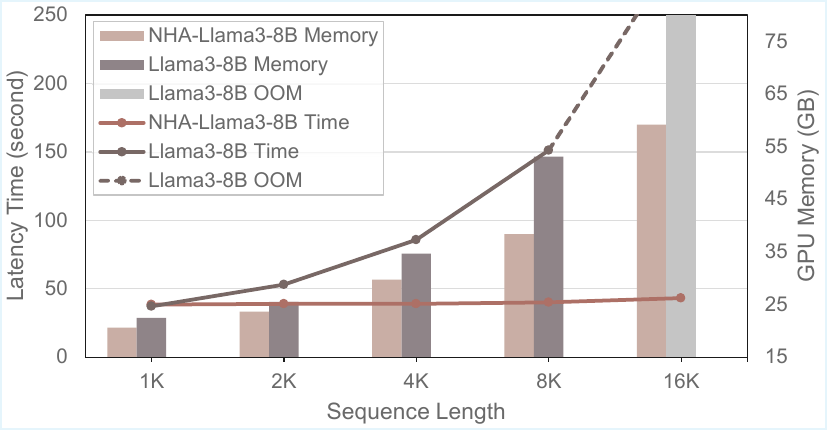}
  \end{center}
  \vspace{-0.8em}
  \caption{\textbf{Efficiency.} Inference latency and memory usage of NHA-Llama3-8B compared with Llama3-8B.}
  \label{fig:efficiency}
  \vspace{-1em}
\end{figure}

To evaluate the efficiency of NHA, we compare the inference speed and GPU memory usage of NHA-Llama3-8B with the original Llama3-8B when generating 1K tokens under varying input. As shown in Figure~\ref{fig:efficiency}, Llama3-8B scales poorly as input length increases, whereas NHA-Llama3-8B shows much slower growth in both metrics, demonstrating superior scalability and efficiency.

\subsection{Ablation Study}
\label{sec:ablation}
In this section, we provide a detailed analysis of the components of NHA to offer insights for the design of future hybrid architectures.

\subsubsection{Inference-Time Architecture Search Exploration}
\label{sec:exploration}

We utilized NHA's flexibility to conduct a training-free architecture search on the FDA benchmark by dynamically adjusting window sizes in Table~\ref{tab:inference_search}. Exploring this vast search space yielded several insights. First, we observed varying sensitivities to global information across layers. Optimizing the placement of Full Attention layers proves as vital as their overall ratio. Highlighting this efficiency, optimizing a 1:8 hybrid baseline by inserting a global window specifically at Layer 11 enabled an NHA with merely 4 Full Attention layers to match a 12-layer baseline. Second, probing the architectural extremes revealed that NHA, when collapsed into a pure Transformer at inference, unexpectedly surpasses a standard Transformer pretrained from scratch. Finally, these zero-shot discovered hybrid configurations demonstrate a downstream value, as utilizing them for specialized pretraining outperforms standard uniform architectures.

\begin{table}[h]
\centering
\adjustbox{max width=\columnwidth}{
\begin{tabular}{llcc}
\toprule
\textbf{Hybrid Ratio} & \textbf{Configuration} & \textbf{Recall Acc.} \\
\midrule
Pure Linear & Block Size 8, Index 64 & 16.71 \\
1:8 & Block Size 8, Index 6	& 44.96 \\
1:8 (Optimized) & Best + Layer 10 & 52.13 \\
1:6 & Block Size 6, Index 5 & 45.05 \\
1:4 & Block Size 4, Index 2 & 51.86 \\
1:3 & Block Size 3, Index 2 & 52.04 \\
1:2 & Block Size 2, Index 0 & 52.68 \\
\midrule
Full Attention NHA & All 24 Layers & \textbf{53.68} \\
Transformer & All 24 Layers & 46.14 \\
\bottomrule
\end{tabular}
}
\caption{\textbf{Design space exploration on FDA.} We evaluate various configurations using a single NHA-24L model. NHA enables a flexible trade-off, bridging the gap between pure linear and full attention models through simple inference-time reconfiguration.}
\label{tab:inference_search}
\end{table}

\subsubsection{Memory and Fusion Ablation}
\vspace{-0.5em}
\begin{table}[h!]
\centering
\small
\renewcommand{\arraystretch}{1.2}
\begin{adjustbox}{width=0.8\columnwidth, center}
\renewcommand{\multirowsetup}{\centering}
\setlength{\tabcolsep}{3pt}
\begin{tabular}{lcc}
\toprule
\textbf{Variant} & \textbf{Recall $\uparrow$} & \textbf{Common $\uparrow$} \\
\midrule
\textbf{NHA (full)} & \textbf{38.60} & 43.09 \\
- w/o Long-Mem & 29.58 & 40.83 \\
- w/o Short-Mem & 36.97 & 41.93 \\
- w/o Token Shift & 35.76 & 41.94 \\
\midrule
- Weighted Sum(F) & 34.06 & 42.69 \\
- Weighted Sum(L) & 33.59 & \textbf{43.12} \\
\bottomrule
\end{tabular}
\end{adjustbox}
\caption{\textbf{Ablation on Memory and Fusion.}}
\label{table:ablation-memory}
\vspace{-0.5em}
\end{table}

We conduct ablations to assess the contributions of NHA’s key components. Removing either long-term or short-term memory leads to clear drops in performance, and eliminating token shift further degrades results by allowing overlap between the two memories. These findings highlight the complementary roles of long- and short-term memory and the necessity of keeping them distinct.

\subsubsection{Alternative Fusion Strategies}
To isolate the effect of the fusion mechanism, we replace NHA's unified softmax with two baselines that follow the common practice in prior ``local+global'' hybrids: first compute \emph{separate} softmax attention over the long-term memory and the short-term memory, obtain their respective outputs, and then combine these outputs via a scalar weight. \textbf{Weighted Sum (F)} uses a fixed coefficient, and \textbf{Weighted Sum (L)} uses a learnable, sequence-dependent coefficient. As shown in Table~\ref{table:ablation-memory}, both alternatives underperform unified softmax, confirming the benefit of joint attention over memories.

\section{Future Work}


The introduction of fixed memory slots and unified long–short memory opens up several promising directions. For instance, \textbf{parameter-efficient finetuning (PEFT)} could be applied to learn slot initial states tailored for downstream tasks. In reasoning scenarios such as \textbf{chain-of-thought (CoT)}, reasoning chains could be selectively compressed into long-term memory, retaining essential information while reducing computational overhead.

\section{Limitations}
NHA achieves strong accuracy and efficiency, but it introduces additional hyperparameters such as slot size and window size that may require careful tuning to fully realize its potential. The current implementation also leaves room for further operator-level optimization, which could further improve deployment efficiency. Moreover, although NHA supports flexible adjustment of window sizes across layers, we primarily consider uniform settings, while more structured strategies such as progressive variation across depth may further enhance adaptability.

\bibliography{custom}

\newpage
\appendix

\section{Answers to Research Questions}
\label{app:ans}

We provide concise answers to the research questions raised in Sec.~\ref{sec:experiments}, with supporting evidence from the corresponding sections.

\paragraph{RQ1: How does NHA's native hybridization design perform against Transformer and other hybrid models?}  
NHA outperforms hybrid baselines on both recall-intensive (Sec.~\ref{sec:recall}), commonsense reasoning tasks (Sec.~\ref{sec:common}) and long-context benchmarks (Sec.~\ref{app:long_context}), confirming stronger overall effectiveness.

\paragraph{RQ2: Can NHA achieve competitive performance to standard Transformers while offering lower cost?}  
Yes. NHA achieves efficiency (Sec.~\ref{sec:ops_efficiency}) and, when applied to pretrained LLMs, closely matches Transformer accuracy with reduced inference time and memory (Sec.~\ref{sec:hybridation-results},~\ref{sec:hybridation-efficiency}).

\paragraph{RQ3: How do NHA's hybrid components contribute, and does unified softmax improve over weighting?}  
Ablations show both memories and token shift are essential, and unified softmax fusion clearly outperforms fixed or learned weighted fusion (Sec.~\ref{sec:ablation}). We also analyze the long-short memory (App.~\ref{appendix:mem}) and conduct ablation study on the slot size and window size (App.~\ref{app:ablation_sw}).

\paragraph{RQ4: Is NHA scalable for production-level LLMs?}  
Yes. Finetuned NHA-Llama and NHA-Qwen demonstrate scalability to billion-scale LLMs with strong accuracy and efficiency gains (Sec.~\ref{sec:hybridation-results}). We scale to Qwen3-30BA3B model and verify the effect of NHA (App.~\ref{app:qwen}). And NHA can utilize fewer full attention layers to get better performance than other hybrid models (App.~\ref{app:attn_layers}).

\section{Related Work and Positioning}

\paragraph{Intra-layer hybrids.}
Existing intra-layer hybrid models typically combine a linear attention mechanism, which maintains long-term information in a matrix memory updated via the outer product of $\bm k$ and $\bm v$, with a local sliding window attention (SWA). Most of these approaches compute the outputs from the linear and local modules \emph{separately} and then merge them via a fixed or learnable weighting. For example, LoLCATs~\citep{zhang2024lolcats} and Liger~\citep{lan2025liger} use fixed, pre-specified fusion coefficients; Infini-attention~\citep{munkhdalai2024leave} employs a learnable but input-independent global coefficient; Griffin~\citep{dong2024hymba} progressively replaces SWA layers with linear–SWA hybrids according to a predefined schedule. In contrast, NHA maintains an explicit $\bm{KV}$ cache for the local window and a compressed slot-based $\bm{KV}$ memory for the long-term context, concatenating the two and applying a \emph{single} softmax over all keys. This yields context-dependent weights that allow the model to dynamically allocate attention between long- and short-term memory without manual fusion rules.

\paragraph{Inter-layer hybrids.}
Inter-layer hybrid architectures mix different layer types within the same model depth. Representative examples include Zamba~\citep{zamba}, Jamba~\citep{jamba}, Samba~\citep{samba}, and GoldFinch~\citep{goldstein2024goldfinch}, which insert Transformer layers at fixed ratios among linear or recurrent layers. Such designs require heterogeneous module types and careful alignment of their hidden representations. By contrast, NHA uses a \emph{single} unified layer design throughout the network. Inter-layer hybridization is achieved simply by adjusting the sliding window size per layer: setting it to zero yields a purely linear layer, while setting it to the full sequence length recovers a Transformer-like layer. This eliminates the need for architectural alignment while still allowing flexible control over the depth-wise allocation of linear and full-attention behavior.

\paragraph{Sharper Contrast to Prior Hybrids}
Many “local+global’’ hybrid models, such as LoLCATs and Infini-attention, follow a two-stage design in which the long-range context is first compressed into global tokens or slots, the local and global attentions are computed independently, and the two outputs are then fused through either a fixed or a learned scalar coefficient. In these approaches, the fusion occurs at the output level, which means the trade-off between global and local information is determined outside the attention computation, is uniform across all tokens for a given query, and is unaffected by the similarity structure between the query and the keys. Gradient flow is also separated: each branch is updated independently based solely on its own attention distribution. Computationally, this design requires running two separate attention operations per query, one for the local branch and one for the global branch.

NHA differs fundamentally in both representation and computation. Long-term context is stored in the same $m\times d$ key–value slot format as local SWA tokens, enabling direct concatenation and a single softmax over all keys. This unified computation produces the global–local allocation $\omega_L$ as part of the attention distribution itself, allowing it to vary per token and per head according to similarity scores with all stored keys. As a result, gradient updates to one memory type are inherently influenced by the logits of the other, creating cross-memory coupling absent in prior designs. Theoretical differences between unified softmax and output-level fusion, including gradient coupling and context-dependent weighting, are analyzed in Sec.~\ref{sec:fusion}. In terms of efficiency, this approach requires only one attention computation over the concatenated set of keys, eliminating the duplicated cost of running separate attentions for each memory type.

Empirically, we complement this analysis with visualization. Appendix~\ref{appendix:mem} presents heatmaps of attention scores for NHA and for representative weighted-fusion baselines, showing that NHA develops position-sensitive long/short allocation patterns that prior methods fail to capture. These visualizations provide further evidence that unified softmax learns richer and more context-dependent allocation dynamics than traditional fusion strategies.

\section{Experiment Details}
\label{app:exp-details}

All experiments were performed on 32 NVIDIA H100 GPUs. Training the 340M-parameter model completed in approximately 2 hours, whereas the 1.3B-parameter model required about 1 day. To ensure reproducibility, we used a fixed random seed of 42 across all training and evaluation procedures. Results are reported from a representative single run, since repeated training confirmed that performance remained consistent across runs.

\subsection{Pretrain Experiment Setup}
Given the scarcity of prior work directly investigating hybrid models that integrate modern linear attention mechanisms such as GLA~\citep{yang2023gated}, GSA~\citep{zhang2024gated}, Gated DeltaNet~\citep{yang2024gated}, and SSMs like Mamba2~\citep{dao2024transformers}, we pretrain all baselines from scratch. Each 340M-parameter model is trained for 15B tokens, and each 1.3B-parameter model for 100B tokens on the SlimPajama~\citep{cerebras2023slimpajama} dataset. All models are optimized with AdamW~\citep{loshchilov2017fixing} (learning rate 3e-4, cosine schedule, weight decay 0.01, gradient clipping 1.0, random seed 42).

\subsection{Finetune Experiment Setup}
\label{appendix:finetune-setup}
We designed our experimental suite to cover a broad range of model sizes, datasets, and baselines, given realistic compute constraints. Our goal was to maximize coverage of scenarios most indicative of real-world performance, while ensuring reproducibility.
\paragraph{Parameter Inheritance}
Given the high parameter similarity between NHA and the standard Transformer, we initialize the corresponding $\bm{Q, K, V}$, and output projection weights in our NHA model directly from the pre-trained checkpoints. For the additional gating parameters introduced by NHA, we adopt a method similar to that in \citep{lan2025liger}, initializing them by applying average pooling to the pretrained $\bm K$-projection weights. For the inter-layer hybrid configuration, we designate one layer in every eight-layer block as a full attention layer, which is achieved by setting its NHA window size to the full sequence length.

\paragraph{Model Hybrid Configuration}
For the 32-layer NHA-Llama3, we designated the first layer in every 8-layer block as a full attention layer, resulting in 4 full attention layers in total. For the 28-layer NHA-Qwen2.5, we designated the second layer in every 7-layer block as a full attention layer, also resulting in 4 full attention layers.

\paragraph{Training Configuration}
We continue the training on the SlimPajama dataset for a total of 10B tokens, divided into two distinct finetuning stages. In the first stage, we freeze the FFN parameters and exclusively finetune the attention layers for 5B tokens, allowing the model to adapt to the new hybrid attention mechanism. Subsequently, in the second stage, we apply LoRA to all model parameters and train for an additional 5B tokens to achieve efficient, full-model finetuning.

\subsection{Large-Scale Hybridization with Qwen3-30B-A3B}
\label{app:qwen}
We further scale up the hybridization experiment to the Qwen3-30B-A3B model. Considering the high computational cost of training such a large model, we adopt a conservative 2:1 hybridization strategy, and train the model on 5B tokens from the SlimPajama dataset.

On standard benchmarks including ARC-Easy, ARC-Challenge \citep{Clark2018ThinkYH}, HellaSwag \citep{zellers2019hellaswag}, LAMBADA \citep{paperno2016lambada}, PIQA \citep{Bisk2020}, and WinoGrande \citep{sakaguchi2019winogrande}, as well as broader evaluations such as GPQA \cite{rein2024gpqa}, IFEval \cite{zhou2023instructionfollowing}, HumanEval \cite{chen2021evaluating}, MathQA \cite{amini2019mathqa} and MMLU \cite{hendrycks2020measuring}. As shown in Table~\ref{tab:qwen} ,the hybridized Qwen3-30B-A3B achieves competitive performance while reducing the reliance on full attention layers. These results confirm the scalability of NHA to tens-of-billions-parameter models.

\subsection{Additional Ablations}

\subsubsection{Ablation on Slot Size and Window Size}
\label{app:ablation_sw}
We jointly vary the number of long-term memory slots $m$ and the sliding window size $w$ to examine their combined effect on NHA's performance.  
All other architectural and training settings are kept identical to the main experiments.  
Table~\ref{table:ablation-slot-window} shows that recall-intensive tasks generally benefit from larger $m$ and moderate $w$, while commonsense reasoning remains stable across a wide range of configurations.  
Extremely small $w$ reduces short-term precision, whereas very large $w$ approaches full attention and erodes efficiency gains.

\begin{table}
\begin{adjustbox}{width=0.8\columnwidth, center}
\small
\renewcommand{\multirowsetup}{\centering}
\setlength{\tabcolsep}{5pt}
\begin{tabular}{ccc c c}
\toprule
\textbf{$m$} & \textbf{$w$} & \textbf{Recall $\uparrow$} & \textbf{Common $\uparrow$} \\
\midrule
64 & 8 & 31.90 & 42.16 \\
64 & 16 & 33.97 & 42.54 \\
32 & 32 & 34.52 & 42.86 \\
64 & 32 & \textbf{38.60} & \textbf{43.09} \\
64 & 64 & 37.83 & 43.06 \\
\bottomrule
\end{tabular}
\end{adjustbox}
\caption{\textbf{Ablation on Slot Size and Window Size.}}
\label{table:ablation-slot-window}
\end{table}

\subsubsection{Ablation on Positional Embedding}

\label{app:ablation-pos}

\begin{table}[h!]
\centering
\small
\renewcommand{\arraystretch}{1.2}
\begin{adjustbox}{width=0.8\columnwidth, center}
\renewcommand{\multirowsetup}{\centering}
\setlength{\tabcolsep}{5pt}
\begin{tabular}{lcc}
\toprule
\textbf{PE Type} & \textbf{Recall $\uparrow$} & \textbf{Common $\uparrow$} \\
\midrule
\textbf{RoPE (S)} & \textbf{38.60} & \textbf{43.09} \\
RoPE (S+L) & 28.46 & 43.01 \\
None & 26.99 & 41.97 \\
Learnable (S) & 30.56 & 41.85 \\
\bottomrule
\end{tabular}
\end{adjustbox}
\caption{\textbf{Ablation on PE strategy.}}
\label{table:ablation-pos}
\end{table}

We evaluate four positional encoding strategies in NHA, focusing on whether and where to apply position embeddings. As shown in Table~\ref{table:ablation-pos}, “S” denotes PE applied to short-term memory and “L” to long-term memory. The results indicate that applying positional encoding solely to short-term memory yields the best performance, as it enhances softmax attention without interfering with the implicit positional encoding already captured by the long-term memory’s recursive update.

\subsection{Relation Between Full Attention Layers and Recall Performance}
\label{app:attn_layers}
\begin{figure}[h!]
    \centering
    \includegraphics[width=0.9\linewidth]{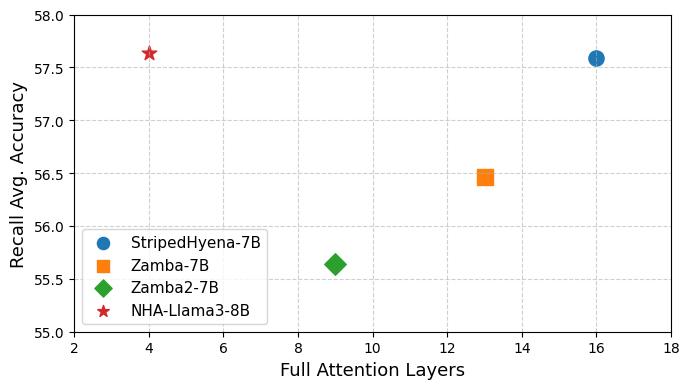}
    \caption{\textbf{Average Recall Accuracy vs. Full Attention Layers.} NHA-Llama3-8B attains the best performance with only 4 full attention layers.}
    \label{fig:hybrid}
\end{figure}

\begin{table*}[t]
\centering
\begin{tabular}{@{}lcccccccc@{}}
\toprule
\textbf{Model} & \textbf{Tokens} & \textbf{ARC-E} & \textbf{ARC-C} & \textbf{Hella.} & \textbf{PIQA} & \textbf{Wino.} & \textbf{MMLU} & \textbf{Avg.} \\ \midrule
Mamba-Llama-3 (12.5\%) & 20B & 70.58 & 43.60 & 71.71 & 75.08 & 59.75 & 49.81 & 61.76 \\
NHA-Llama-3 (12.5\%) & \textbf{10B} & \textbf{80.76} & \textbf{52.47} & \textbf{78.93} & \textbf{79.71} & \textbf{72.85} & \textbf{60.21} & \textbf{70.82} \\ \bottomrule
\end{tabular}
\caption{Zero-shot performance comparison. Both models are distilled from Llama-3 with a 12.5\% hybrid ratio.}
\label{tab:mamba_comparison}
\end{table*}

\subsection{Advantage over Mamba-H: Structural Compatibility}
\label{sec:appendix_mamba_comparison}

We further demonstrate NHA's superior adaptation efficiency by comparing it against Mamba-Llama-3 \cite{wang2024mamba}. Because NHA natively unifies linear and standard attention, it treats Full Attention simply as a special case with a full-length window. This structural compatibility enables highly efficient parameter inheritance when distilling from dense Transformer models like Llama-3. As shown in Table \ref{tab:mamba_comparison}, when configuring both models with a 12.5\% hybrid ratio, NHA-Llama-3 achieves a significantly higher average zero-shot score across diverse benchmarks. Remarkably, NHA achieves this $+9.06$ performance gap while consuming only half the training tokens, validating its exceptional cost-effectiveness for downstream adaptation.

\section{Gradient Coupling Analysis}
\label{app:grad-coupling}

In prior hybrids, the fusion coefficient between long- and short-term outputs is typically fixed or predicted only from the current input, without access to the similarity distribution across both memories. As a result, such coefficients cannot in general reproduce $\omega_L$, which reflects both the current query and the aggregate interactions with all stored tokens. Because the weighting is computed jointly over all tokens in both memories, the model can express fine-grained, query-dependent trade-offs that static or input-only fusion coefficients cannot represent.

\noindent \textbf{Gradient coupling.}
Let $Z_L = \sum_{i\in\text{long}}\exp(\bm q_t^\top \bm k_i)$ and $Z_S = \sum_{j\in\text{short}}\exp(\bm q_t^\top \bm k_j)$. Define the unified attention distribution over all keys as $p_{\text{uni}}(u) = \frac{\exp(\bm q_t^\top \bm k_u)}{Z_L+Z_S}$. Differentiating Eq.~\ref{eq:context-fusion} with respect to the logits $\ell_u = \bm q_t^\top \bm k_u$ yields:

\begin{align}
    \frac{\partial \omega_L}{\partial \ell_i} &= p_{\text{uni}}(i)\,(1-\omega_L), \quad i \in \text{long}, \\
    \frac{\partial \omega_L}{\partial \ell_j} &= -p_{\text{uni}}(j)\,\omega_L, \quad j \in \text{short}.
\end{align}

These expressions show that the gradient adjusting the long/short allocation is modulated by \emph{both} memories: changes in long-term logits depend on the total short-term mass $1-\omega_L$, and vice versa. This creates a natural cross-memory coupling that aligns both memories in the same similarity space.

By contrast, in simple weighted fusion where the coefficient $\alpha\_t$ is applied after computing two independent softmaxes, $\frac{\partial \alpha_t}{\partial \ell_u} = 0$ for any token $u$ not used to compute $\alpha_t$. As a result, logits in the other memory type receive no gradient signal for adjusting the allocation, leading to weaker co-adaptation between memories.

\section{Memory Analysis}
\label{appendix:mem}
\begin{figure*}[h!]
    \centering
    \includegraphics[width=1.0\linewidth]{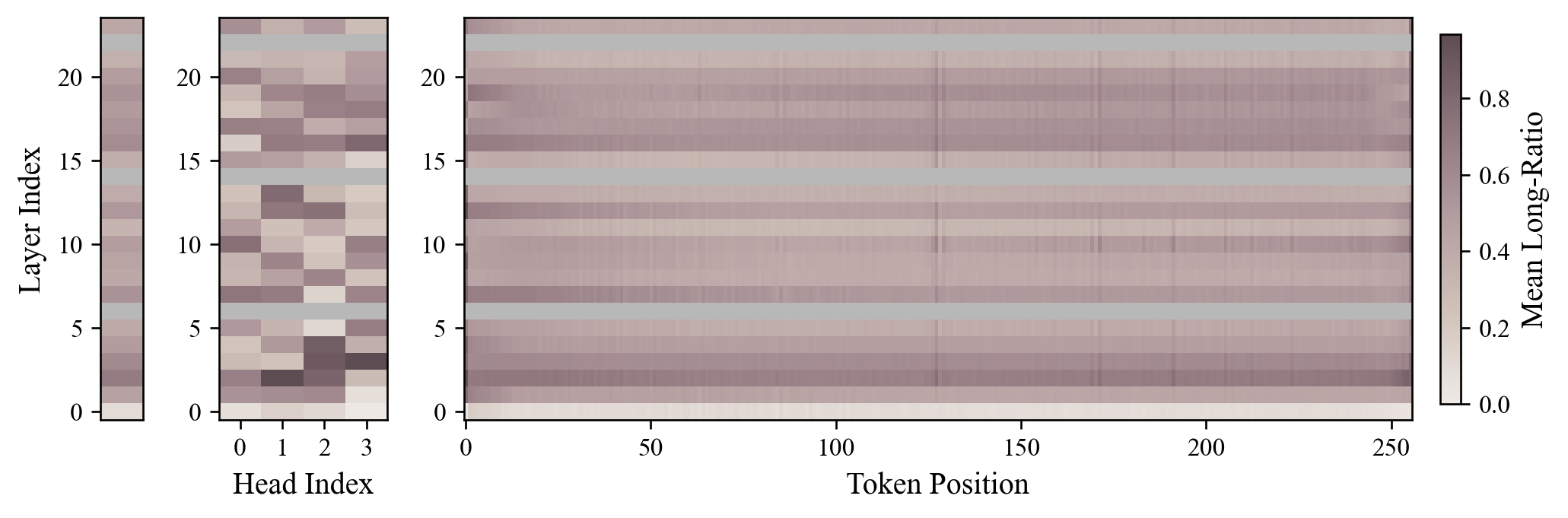}
    \caption{\textbf{Visualization of Long-Term Memory Usage.} From left to right: (1) layer-wise mean long-ratio averaged over all heads and positions; (2) average long-ratio per layer–head pair aggregated over sequence positions; (3) average long-ratio per layer–position pair aggregated over attention heads. Gray-shaded layers indicate hybrid Transformer layers in the inter-layer configuration.}
    \label{fig:nha_long_ratio}
\end{figure*}

\begin{figure*}[h!]
    \centering
    \includegraphics[width=1.0\linewidth]{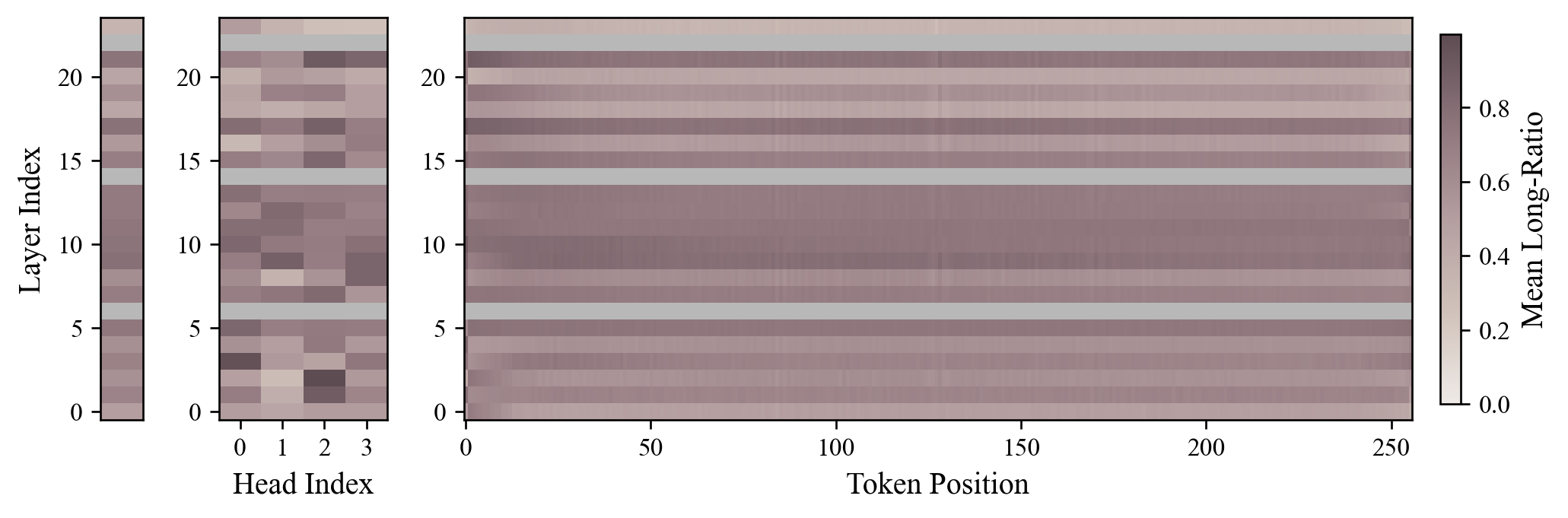}
    \caption{\textbf{Visualization of Long-Term Memory Usage.} Across layers of the input-projection fusion variant.}
    \label{fig:add_long_ratio}
\end{figure*}

We analyze the distribution of attention between long- and short-term memory by visualizing the average long-ratio across layers, heads, and positions (Figure~\ref{fig:nha_long_ratio}). The results show that different layers and heads exhibit distinct preferences, indicating a specialization in memory usage. Positional patterns are also evident: early tokens before the sliding window is filled rely more on long-term memory, while later tokens increasingly attend to long-term slots for retrieving distant information.

For comparison, we examined a variant where the fusion weight between long- and short-term memory is obtained by applying a learnable linear projection to the current input (Figure~\ref{fig:add_long_ratio}). This variant produces a distribution that is overall more uniform across sequence positions, showing little variation between early and late tokens. In contrast, NHA exhibits a clear increase in long-term memory usage toward later positions, reflecting its ability to adapt to the growing need for retrieving distant information. This highlights that unified softmax not only enables per-token, context-dependent weighting but also captures position-sensitive patterns. This adaptive positional trend is difficult to obtain with weighted-sum fusion, where the long–short memory trade-off is externally parameterized rather than jointly learned within the attention distribution.

\section{Datasets}
We pretrain our models on the SlimPajama dataset. For the 340M variant, we use a 15B-token sample, while the 1.3B variant is trained on a 100B-token sample. SlimPajama \cite{cerebras2023slimpajama} is an English-language, high-quality subset of RedPajama that includes Common Crawl, Wikipedia, books, and GitHub code. It is cleaned, deduplicated, and optimized for large-scale model training.

For standard language understanding, we evaluate on the following English benchmarks: WikiText (62 test samples) \cite{merity2016pointer}, derived from Wikipedia articles authored by global volunteers; LAMBADA (5153) \cite{paperno2016lambada}, sourced from narrative books; ARC-Easy (2376) and ARC-Challenge (1172) \cite{Clark2018ThinkYH}, consisting of science exam questions written by educators; HellaSwag (10003) \cite{zellers2019hellaswag}, constructed from activity descriptions such as WikiHow; PiQA (3084) \cite{Bisk2020}, crowdsourced for physical commonsense reasoning; and WinoGrande (1267) \cite{sakaguchi2019winogrande}, a large-scale pronoun resolution dataset generated through crowdsourcing.

To assess recall-intensive abilities, we adopt FDA (1102 test samples) \cite{arora2024simple, arora2023language}, containing annotated medical device submissions; SWDE (1111) \cite{arora2024simple, arora2023language, lockard-etal-2019-openceres}, curated from movie and university websites; SQuAD (2984) \cite{rajpurkar2018know}, based on Wikipedia question answering; Natural Questions (3157) \cite{kwiatkowski2019natural}, sourced from Google search queries; TriviaQA (1688) \cite{joshi2017triviaqa}, composed of trivia-style questions with web evidence; and DROP (2087) \cite{dua2019drop}, Wikipedia passages requiring discrete reasoning.

All datasets are in English, publicly available, and released by their original creators. They are used strictly under their intended purposes and licenses, without modification or derivative dataset creation.

\section{Use of AI Assistants}
AI assistants were employed solely for improving the clarity and readability of the manuscript through minor language polishing. They were not involved in study design, theoretical development, experiments, data analysis, or any other substantive aspect of this research.

\end{document}